\documentclass[]{article}
\pdfoutput=1
\usepackage{proceed2e}
\usepackage{times}
\usepackage{epsfig}
\usepackage{graphicx}
\usepackage{amsmath}
\usepackage{amssymb}
\usepackage{subfigure}

\title{Bregman Distance to L1 Regularized
Logistic Regression}

\author{
Mithun Das Gupta  \\
Epson Research and Development, Inc. \\
        2580 Orchard Parkway, Suite 225 \\
        San Jose, CA 95131. \\
\texttt{mdasgupta@erd.epson.com} \\
\And
Thomas S. Huang \\
Dept. Electrical and Computer Engg. \\
Beckman Inst. of Advance Science and Tech. \\
University of Illinois, Urbana Champaign \\
\texttt{huang@ifp.uiuc.edu} \\
}

\begin{document}

\maketitle

\begin{abstract}
In this work we investigate the relationship between Bregman
distances and regularized Logistic Regression model. We present a
detailed study of Bregman Distance minimization, a family of
generalized entropy measures associated with convex functions. We
convert the L1-regularized logistic regression into this more
general framework and propose a primal-dual method based algorithm
for learning the parameters. We pose L1-regularized logistic
regression into Bregman distance minimization and then apply
non-linear constrained optimization techniques to estimate the
parameters of the logistic model.
\end{abstract}

\section{Introduction}
We study the problem of regularized logistic regression as proposed
by~\cite{Collins02} and~\cite{LaffertyDella97}. $L1$
regularization has been studied extensively during recent years due
to the sparsity of the classifiers obtained by such
regularization~\cite{Ng04}. 
The objective function in the $L1$-regularized LRP
(Eqn.~\ref{Eqn:L1REg}) is convex, but not differentiable
(specifically, when any of the weights is zero), so solving it is
more of a computational challenge than solving the $L2$-regularized
LRP. Despite the additional computational challenge posed by
$L1$-regularized logistic regression, compared to $L2$-regularized
logistic regression, interest in its use has been growing. The main
motivation is that $L1$-regularized LR typically yields a sparse
vector $\boldsymbol{\lambda}$, i.e., $\boldsymbol{\lambda}$
typically has relatively few nonzero coefficients. (In contrast,
$L2$-regularized LR typically yields $\boldsymbol{\lambda}$ with all
coefficients nonzero.) When ${\lambda}_j$ = 0, the associated
logistic model does not use the jth component of the feature vector,
so sparse $\boldsymbol{\lambda}$ corresponds to a logistic model
that uses only a few of the features, i.e., components of the
feature vector. Indeed, we can think of a sparse
$\boldsymbol{\lambda}$ as a selection of the relevant or important
features (i.e., those associated with nonzero ${\lambda}_j$), as
well as the choice of the intercept value and weights (for the
selected features). A logistic model with sparse
$\boldsymbol{\lambda}$ is, in a sense, simpler or more parsimonious
than one with non-sparse $\boldsymbol{\lambda}$. It is not
surprising that $L1$-regularized LR can outperform $L2$-regularized
LR, especially when the number of observations is smaller than the
number of features.

Our work is based directly on the general setting
of~\cite{LaffertyDella97} in which one attempts to solve
optimization problems based on general Bregman distances. They
proposed the iterative scaling algorithm for minimizing such
divergences through the use of auxiliary functions. Our work builds
on several previous works which have compared divergence approaches
to logistic regression. We closely follow the work
by~\cite{Collins02} who propose a new category of parallel
and sequential algorithms for boosting and logistic regression based
on Bregman distance minimization. They are one of the first to
connect the fields of regression and generalized divergences, but as
such unconstrained logistic parameter is unreliable for large
problems and hence we take up this study to tie constrained
optimization to the existing work.

Most of the work related to connecting the idea of Bregman distance
and logistic regression minimize the unconstrained auxiliary
function at each step. In this work we pose the problem with box or
$L1$ constraints due to the favorable properties of $L1$
regularization for cases with large dimensions but relatively fewer
number of training data points.

\section{Logistic Regression}

Let $\mathcal{S}=\langle(x_1,y1),\dots,(x_m,y_m)\rangle$ be a set of
training examples where each instance $x_i$ belongs to a domain or
instance space $\boldsymbol{\chi}$ , and each label $y_i\in
\{-1,+1\}$.

We assume that we are given a set of real-valued functions on
$\boldsymbol{\chi}$, denoted by $h_i$ where $i=\{1,2,\dots,n\}$.
Following convention in the Maximum-Entropy literature, we call
these functions features; in the boosting literature, these would be
called weak or base hypotheses. Note that, in the terminology of the
latter literature, these features correspond to the entire space of
base hypotheses rather than merely the base hypotheses that were
previously found by the weak learner. We study the problem of
approximating the $y_i$'s using a linear combination of features.
That is, we are interested in the problem of finding a vector of
parameters $\boldsymbol{\lambda}\in \mathbb{R}^n$ such that
$f_{\lambda}(x_i)=\sum_{j=1}^n\lambda_jh_j(x_i)$ is a good
approximation of $y_i$.

For classification problems, it is natural to try to match the sign
of $f_{\boldsymbol{\lambda}}(x_i)$ to $y_i$, that is, to attempt to
minimize

\begin{equation}\label{Eqn:classificationLoss}
    \sum_{j=1}^nI_{\mathbb{[}y_if_{\boldsymbol{\lambda}}(x_i)\le 0\mathbb{]}}
\end{equation}

where $I_{\{c\}}=1$ whenever \{c\} is $true$. This form of loss is
intractable for in its most general form and so some other
non-negative loss function is minimized which closely resembles the
above loss.

In the logistic regression framework we use the estimate
\begin{equation}\label{Eqn:LogisticRegression}
    \mathrm{P}\{y=+1|x\}=\frac{1}{1+\exp(-f_{\boldsymbol{\lambda}}(x))}
\end{equation}
and the log-loss for this model is defined as
\begin{equation}\label{Eqn:LogLoss}
    \ell(\mathbf{x},\mathbf{y})=\sum_{j=1}^m\ln(1+exp(-y_if_{\boldsymbol{\lambda}}(x_i)))
\end{equation}

This is the loss function for the unconstrained minimization
problem. But as pointed out earlier regularized loss functions are
effective for most practical cases and hence we would try to pose
the optimization problem with the regularized loss function. The
regularized loss function can now be written as
\begin{equation}\label{Eqn:L1REg}
    \ell(\mathbf{x},\mathbf{y})=\sum_{j=1}^m\ln(1+exp(-y_if_{\boldsymbol{\lambda}}(x_i)))
    + R(\boldsymbol{\lambda})
\end{equation}
where $R(\boldsymbol{\lambda})$ is the regularization function and
can have different forms depending on the regularization method. For
$L1$ regularization the function $R$ is defined as
$\alpha|\boldsymbol{\lambda}|_1$.

\section{Bregman Distance}\label{Sec:BMDistance}
Let $F:\Delta\rightarrow \mathbb{R}$ be a continuously
differentiable and strictly convex function defined on a closed
convex set $\Delta \subseteq \mathbb{R}_+^r$. The Bregman distance
associated with $F$ is defined for $\mathbf{p},\mathbf{q}\in \Delta$
to be

\begin{equation}\label{Eqn:BregmanDistance}
    B_F(\mathbf{p}\;\|\;\mathbf{q})\doteq
    F(\mathbf{p})-F(\mathbf{q})-\nabla
    F(\mathbf{q})\cdot(\mathbf{p}-\mathbf{q})
\end{equation}

For instance when

\begin{equation*}
    F(\mathbf{p})=\sum_{i=1}^rp_i\ln p_i
\end{equation*}

$B_F$ is the unnormalized relative entropy, defined as $D_U$

\begin{equation*}
    D_U(\mathbf{p}\;\|\;\mathbf{q})=\sum_{i=1}^r\left(
                                              \begin{array}{c}
                                                p_i\ln\left(
                                                        \begin{array}{c}
                                                          \frac{p_i}{q_i} \\
                                                        \end{array}
                                                      \right)
                                                +q_i-p_i \\
                                              \end{array}
                                            \right)
\end{equation*}

\begin{figure}[ht]
\begin{center}
\includegraphics[width=5cm]{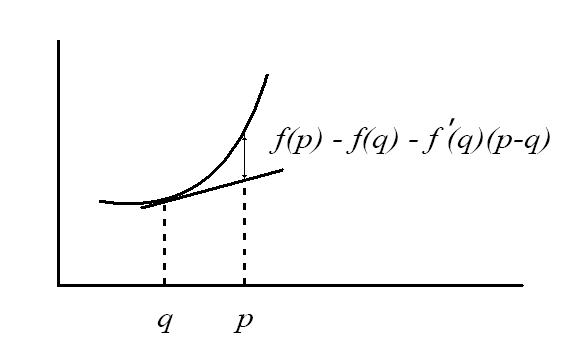}
\caption{The Bregman distance $B_f(p\;\|\; q)$ is an indication of
the increase in $f(p)$ over $f(q)$ above linear growth with slope
$f'(q)$.} \label{Fig:BregmanDist}
\end{center}
\end{figure}

A graphical representation of Bregman distance as a measure of
convexity is shown in Fig.~\ref{Fig:BregmanDist}.

The distances $B_F$ were introduced in by Bregman~\cite{Bregman67}
along with an iterative algorithm for minimizing $B_F$ subject to
linear constraints.
Bregman distances have been used earlier by numerous authors to pose
problems as generalized divergences.~\cite{Dhillon05} used
such divergences for generalized nonnegative matrix
approximations.~\cite{Banerjee04} used them for clustering
applications. Other divergence minimization approaches have been
tried for data mining and information retrieval. The concept of
posing numerous problems of density estimation as KL divergence
minimization problem has been long studied. It can be shown that KL
divergence is a specialized case of Bregman divergence and hence the
comprehensive success of such methods warrants a better
investigation of Bregman divergence itself.

To develop the rest of this work we need a few definitions. Let
$\Delta\subset \mathbb{R}^r$ and let $F:\Delta\rightarrow
\mathbb{R}$ be a real valued function. We assume that $\Delta$ is a
closed convex set, and that $F$ is strictly convex and $C^1$ on the
interior of $\Delta$.

{\bf Definition 1} {\it For $\mathbf{v}\in \mathbb{R}^r$ and
$\mathbf{q}\in \Delta$ the Legendre Transform $
\mathcal{L}_F(\mathbf{v},\mathbf{q})$ is defined as }

\begin{equation*}
    \mathcal{L}_F(\mathbf{v},\mathbf{q})=\arg \displaystyle\min_{\mathbf{p}\in
    \Delta} B_F(\mathbf{p}\;\|\;\mathbf{q}) + \mathbf{v}\cdot\mathbf{q}
\end{equation*}

{\bf Lemma 1} {\it The mapping
${\mathbf{v},\mathbf{q}}\mapsto\mathcal{L}_F(\mathbf{v},\mathbf{q})$
defines a smooth action of $\mathbb{R}^r$ on $\Delta$ by}
\begin{equation*}
    \mathcal{L}_F(\mathbf{v},\mathcal{L}_F(\mathbf{w},\mathbf{q}))=\mathcal{L}_F((\mathbf{v}+\mathbf{w}),\mathbf{q}).
\end{equation*}


The optimization problem which we consider is the following: let $A$
be an $n\times r$ matrix of linear constraints on $\mathbf{p}\in
\Delta$. Let $\mathbf{q}_0\in \Delta$ be a {\it default
distribution}, chosen such that $\nabla F(\mathbf{q}_0)=0$. Finally,
let $\tilde{\mathbf{p}}\in \Delta$ be given, which is considered the
{\it empirical distribution}, since it typically arises from a set
of training samples that determine the linear constraints.

We now define $\mathcal{P}(A,\tilde{\mathbf{p}})$ and
$\mathcal{Q}(A,\mathbf{q}_0)$ as

\begin{eqnarray*}
  \mathcal{P}(A,\tilde{\mathbf{p}}) &=& \{\mathbf{p}\in \Delta | A\mathbf{p}=A\tilde{\mathbf{p}} \} \\
\mathcal{Q}(A,\mathbf{q}_0) &=& \{\mathbf{q} \in \Delta |
\mathbf{q}=\mathcal{L}_F((\boldsymbol{\lambda}^TA),\mathbf{q}_0),\boldsymbol{\lambda}
\in \mathbb{R}^n \}
\end{eqnarray*}

The following well-known theorem~\cite{LaffertyDella97} establishes
the duality between the two natural projections of
$B_F(\mathbf{p}\;\|\;\mathbf{q})$ with respect to the families
$\mathcal{P}(A,\tilde{\mathbf{p}})$ and
$\mathcal{Q}(A,\mathbf{q}_0)$

{\bf Theorem 1} {\it Suppose
$B_F(\tilde{\mathbf{p}}\;\|\;\mathbf{q})<\infty$ and let
$\bar{\mathcal{Q}}(A,\mathbf{q}_0)=\mathrm{cl}(\mathcal{Q}(A,\mathbf{q}_0))$.
Then there exists a unique $\mathbf{q}_{\star} \in \Delta$ such
that}

\begin{enumerate}
\item[1.]$\mathbf{q}_{\star} \in \mathcal{P}(A,\tilde{\mathbf{p}}) \cap \mathcal{Q}(A,\mathbf{q}_0)$
\item[2.]$B_F(\mathbf{p}\;\|\;\mathbf{q})=B_F(\mathbf{p}\;\|\;\mathbf{q}_{\star})+B_F(\mathbf{q}_{\star}\;\|\;\mathbf{q})$
for any $\mathbf{p}\in \mathcal{P}(A,\tilde{\mathbf{p}})$ and
$\mathbf{q}\in \bar{\mathcal{Q}}(A,\mathbf{q}_0)$
\item[3.]$\mathbf{q}_{\star}=\arg \displaystyle\min_{\mathbf{q}\in
    \bar{\mathcal{Q}}}B_F(\tilde{\mathbf{p}}\;\|\;\mathbf{q})$
\item[4.]$\mathbf{q}_{\star}=\arg \displaystyle\min_{\mathbf{p}\in
    \bar{\mathcal{P}}}B_F({\mathbf{p}}\;\|\;\mathbf{q}_0)$
\end{enumerate}
{\it Moreover, any of these four properties determines
$\mathbf{q}_{\star}$ uniquely.}

Note that since we have defined $\nabla F(\mathbf{q}_0)=0$, $\arg
\displaystyle\min_{\mathbf{p}\in\bar{\mathcal{P}}}B_F({\mathbf{p}}\;\|\;\mathbf{q}_0)=\arg
\displaystyle\min_{\mathbf{p}\in\bar{\mathcal{P}}}F(\mathbf{p})$.
Property $2$. is called the {\it Pythagorean property} since it
resembles the Pythagorean theorem if we imagine that
$B_F(\mathbf{p}\;\|\;\mathbf{q})$ is the square of Euclidean
distance and $(\mathbf{p},\mathbf{q}_{\star},\mathbf{q})$ are the
vertices of a right triangle.


\section{Bregman Distance to Logistic Regression} In this section we study
the minimization problem as mentioned in the previous section. By
unconstrained we mean that the parameters $\lambda\in \mathbb{R}^n$
are free. We pose the logistic regression problem in the Bergman
distance framework which was developed by Collins and
Schapire~\cite{Collins02}.

The key idea is to write the function $F(\mathbf{p})$ as
\begin{equation}\label{Eqn:LogFunc}
    F(\mathbf{p})=\sum_{i=1}^m p_i\ln p_i + (1-p_i)\ln(1-p_i)
\end{equation}
The resulting Bergman distance is

\begin{equation}\label{Eqn:LogisticBregman}
    D_B(\mathbf{p}\;\|\;\mathbf{q})=\sum_{i=1}^m p_i\ln\frac{p_i}{q_i}	+ (1-p_i)\ln\frac{1-p_i}{1-q_i}
\end{equation}

For this choice of $F$ the Legendre transform is found to be

\begin{equation}\label{Eqn:Legendre}
    \mathcal{L}_F(v,q)_i=\frac{q_ie^{-v_i}}{1-q_i+q_ie^{-v_i}}
\end{equation}

Now we define the constraint matrix $A$ as $A_{ji}=y_ih_j(x_i)$ from
which we get
$v_i=(\boldsymbol{\lambda}^TA)i=\sum_{j=1}^n\lambda_jy_ih_j(x_i)$

Now, if we put $\mathbf{q_0}=(1/2)\mathbf{1}$ into
eqn.~\ref{Eqn:Legendre} we get the logistic probability
eqn.~\ref{Eqn:LogisticRegression}.

Also note that
\begin{equation}\label{Eqn:DB0q}
    D_B(\mathbf{0}\;\|\;\mathbf{q})=-\sum_{i=1}^m\ln(1-q_i)
\end{equation}

which gives

\begin{eqnarray}\label{EQN:DB2logloss}
    \ell(\mathbf{x},\mathbf{y})&=&\sum_{i=1}^m\ln(1+e^{(-y_if_{\boldsymbol{\lambda}}(x_i)})\\ \nonumber
    &=&D_B(\mathbf{0}\|\mathcal{L}_F(\boldsymbol{\lambda}^T
    A,\mathbf{q_0}))
\end{eqnarray}

where $f_{\boldsymbol{\lambda}}(x_i)=\sum_{j=1}^n\lambda_jh_j(x_i)$

 Finally, we can write the equivalent optimization problem as

\begin{eqnarray}
\nonumber
  \displaystyle\min_{\mathbf{q}\in\bar{\mathcal{Q}}} &&	 D_B(\mathbf{0}\;\|\;\mathbf{q})\\
   st && A\mathbf{q}=0
\end{eqnarray}

where as before $\bar{\mathcal{Q}}=\mathrm{cl}(\mathcal{Q})$, where

\begin{equation*}
Q=\left\{
    \begin{array}{c}
      \mathbf{q}\in \Delta : q_i = \sigma\left(
                               \begin{array}{c}
                                 \sum_{j=1}^n\lambda_j y_i h_j (x_i) \\
                               \end{array}
                             \right), \boldsymbol{\lambda}\in \mathbb{R}^n \\
    \end{array}
  \right\}
\end{equation*}

where $\sigma(x)=(1+e^x)^{-1}$ is the Sigmoid function. For our
choice of $\boldsymbol{q_0}=(1/2) \boldsymbol{1}$ we have
$\mathcal{L}_F(\boldsymbol{v},\mathbf{q_0})_i=\sigma(v_i)$ as shown
in Eqn.~\ref{Eqn:Legendre}. Also, since each of the elements of
$\boldsymbol{q}$ is Sigmoid function output, therefore, $\Delta \in
[0,1]^m$.

The key points to note in this derivation are
\begin{enumerate}
\item[a.]$\tilde{{\mathbf{p}}}\equiv 0$
\item[b.]$\boldsymbol{\lambda}\in \mathbb{R}^n$
\end{enumerate}

The implication of the point (a.) above is that the constraints are
homogenous. This is a strong assumption on the constraints. It so
turns out that we can relax this constraint only when we put some
additional constraints on the free parameter $\boldsymbol{\lambda}$.
This points to a regularized scheme, where the first constraint is
relaxed on the cost of putting some additional constraints on the
second condition. We redefine the set $\mathcal{Q}$ as

\begin{equation*}
Q=\{\mathbf{q} : q_i = \sigma(\sum_{j=1}^n\lambda_j y_i h_j (x_i)) ,
\boldsymbol{\lambda}\in \mathbb{R}^n,\|\boldsymbol{\lambda}\|_1\le c
\}
\end{equation*}

We consider supervised learning in settings where there are many
input features, but where there is a small subset of the features
that is sufficient to approximate the target concept well. In
supervised learning settings with many input features, over-fitting
is usually a potential problem unless there is ample training data.
For example, it is well known that for un-regularized discriminative
models fit via training error minimization, sample complexity (i.e.,
the number of training examples needed to learn ``well'') grows
linearly with the VC dimension~\cite{VapnikChervo74}. Further, the
VC dimension for most models grows about linearly in the number of
parameters~\cite{Vapnik82}, which typically grows at least linearly
in the number of input features. Thus, unless the training set size
is large relative to the dimension of the input, some special
mechanism, such as regularization, which encourages the fitted
parameters to be small is usually needed to prevent over-fitting.

Once we have defined our optimization problem our aim is to find a
sequence of
$q_k=\mathcal{L}_F(\boldsymbol{\lambda}_k^TA,\mathbf{q_0})$ which
minimizes our cost function, all the while remaining feasible to the
additional regularization constraint
$\;\|\;\boldsymbol{\lambda}\;\|\;_1\le c$.

\section{Auxiliary Function} The idea of auxiliary functions was
proposed by Della Pietra et al.~\cite{LaffertyDella97}. The
idea is analogous to EM algorithm and tries to bound the error for
two iterations. Since we are dealing with distances which are
defined to be positive, so the quantity
$\|d_{t+1}-d_t\|=-(d_{t+1}-d_t)$ for strict descent, which can be
minimized iteratively, till convergence is achieved.

{\bf Definition 2 \hspace{0.2cm}}{\it For a linear constraint matrix
$A$, if $\boldsymbol{\lambda}\in \mathbb{R}^n$. A function
$\mathcal{A}:\mathbb{R}^n\times \Delta \rightarrow \mathbb{R}$ is an
auxiliary function for $L(q)=-B_F(\tilde{p}\;\|\;q)$ if}
\begin{enumerate}
\item[1.] For all $q\in \Delta$ and $\boldsymbol{\lambda}\in \mathbb{R}^n$

$L(\mathcal{L}_F(\boldsymbol{\lambda}^TA,\;q))\ge L(q)+
\mathcal{A}(\boldsymbol{\lambda},q)$
\item[2.] $\mathcal{A}(\boldsymbol{\lambda},q)$ is continuous in $q\in
\Delta$ and $C^1$ in $\boldsymbol{\lambda}\in \mathbb{R}^n$ with
$\mathcal{A}(0,q)=0$ and

\begin{equation*}
    \frac{d}{dt}|_{t=0}\mathcal{A}(t\boldsymbol{\lambda},q)=\frac{d}{dt}|_{t=0}L(\mathcal{L}_F(((t\boldsymbol{\lambda})^TA),q))
\end{equation*}

\item[3.] If $\boldsymbol{\lambda}=0$ is a minima of
$\mathcal{A}(\boldsymbol{\lambda},q)$, then $q^TA=p_0^TA$.
\end{enumerate}

{\bf Theorem 2\hspace{0.4cm}}{\it Suppose $q^k$ is any sequence in
$\Delta$ with $q^0 = q_0$ and
$q^{k+1}=\mathcal{L}_F(\boldsymbol{\lambda}^TA,q)$ where
$\boldsymbol{\lambda}\in \mathbb{R}^n$ satisfies}
\begin{eqnarray*}
  \mathcal{A}(\boldsymbol{\lambda}_k,q^k)&=&\displaystyle\sup_{\boldsymbol{\lambda}}\mathcal{A}(\boldsymbol{\lambda},q^k)
\end{eqnarray*}
{\it Then $L(q^k)$ increases monotonically to
$\displaystyle\max_{q\in \bar{\mathcal{Q}}}L(q)$ and $q^k$ converges
to the distribution $q_{\star}=\arg \displaystyle\max_{q\in
\bar{\mathcal{Q}}}L(q)$.}

The proof of this theorem is elucidated in Della Pietra et
al.~\cite{LaffertyDella97}. We will mention the three lemmas on
which the proof is based. Once the lemmas have been proved the proof
for the theorem can be drawn simply from them. The three lemmas are
\begin{enumerate}
\item[1.] If $m\in \Delta$ is a cluster point of $\boldsymbol{q}^{(k)}$,
then $\mathcal{A}(\boldsymbol{\lambda},\boldsymbol{q}^{(k)})\le 0$
for all $\boldsymbol{\lambda}\in \mathbb{R}^n$.
\item[2.]If $m\in \Delta$ is a cluster point of $\boldsymbol{q}^{(k)}$, then
$\frac{d}{dt}|_{t=0}L(\mathcal{L}_F(t\boldsymbol{\lambda}^TA,\boldsymbol{q}^{(k)}))=0$
for all $\boldsymbol{\lambda}\in \mathbb{R}^n$.
\item[3.]Suppose $\{\boldsymbol{q}^{(k)}\}$ is any sequence with only one cluster
point $\boldsymbol{q}_{\star}$ . Then $\boldsymbol{q}^{(k)}$
converges to $\boldsymbol{q}_{\star}$.
\end{enumerate}

\section{Constrained Bregman Distance Minimization} Once we have
shown the analogy between logistic regression and Bregman distances,
we can proceed to find a suitable auxiliary function for our
problem. One key observation is that we can write $q_{k+1}$ as a
simple function of $q_{k}$ as follows
\begin{eqnarray*}
    \boldsymbol{q}_{k+1}&=&\mathcal{L}_F((\boldsymbol{\lambda}_k+\boldsymbol{\delta}_k)^TA,\;\boldsymbol{q_0})\\
    &=& \mathcal{L}_F(\boldsymbol{\delta}_k^TA,\;\mathcal{L}_F(\boldsymbol{\lambda}_k,\;\boldsymbol{q_0})) \\
    &=& \mathcal{L}_F(\boldsymbol{\delta}_k^TA,\;\boldsymbol{q}_k) \\
\end{eqnarray*}
Let us denote $\mathbf{v}=\boldsymbol{\delta}_k^TA$, hence we can
write $\boldsymbol{q}^{k+1}= \mathcal{L}_F(v,\boldsymbol{q}_k)$.
Now, from Eqn.~\ref{Eqn:DB0q}, we can write

\begin{eqnarray*}
  D_B(0\|\boldsymbol{q}^{k+1})-D_B(0\|\boldsymbol{q}^{k}) &=&
  \sum_{i=1}^m\ln(1-q_i+q_ie^{-v_i})\\
  &\le&\sum_{i=1}^mq_i(e^{-v_i}-1)
\end{eqnarray*}

Substituting, $(\boldsymbol{\delta}^TA)_i=\mathbf{v}_i$, we define
our auxiliary function as

\begin{equation}\label{EQN:AuxFunc}
    \mathcal{A}(\boldsymbol{\delta},\boldsymbol{q})=\sum_{i=0}^mq_i(e^{-(\boldsymbol{\delta}^TA)_i}-1)
\end{equation}

It can be easily verified that the above choice of auxiliary
function satisfies the conditions mentioned in Def 2. Now we need to
find a sequence of $\{\delta^k\}\rightarrow 0$ for which
$\mathcal{A}(\boldsymbol{\delta},\boldsymbol{q})\le 0$ and
$\mathcal{A}(\boldsymbol{\delta},\boldsymbol{q})\rightarrow 0$
monotonically.

\section{Algorithm}

{\bf Assumptions:} $F:\Delta \rightarrow \mathbb{R}$, such that
$\{q\in \Delta: B_F(0\;\|\;\boldsymbol{q})\le c\}$ where $c<\infty$.

{\bf Parameters:} $\Delta \in [0,1]^m$, $F$ satisfying assumptions
in part 1, and
$\boldsymbol{q}_0=(1/2)\mathbf{1}$. 

{\bf Input:} Constraint matrix $A\in [-1,1]^{n\times m}$, where
$A_{ji}=y_ih_j(x_i)$, and $\sum_{j=1}^n|A_{ji}|\le
1$.

{\bf Output:} Denote
$\mathcal{L}_F(\boldsymbol{\lambda}_t^TA,\boldsymbol{q}_0)$ as
$\mathcal{L}_F^{\boldsymbol{\lambda}_t}$. Generate a sequence of
$\boldsymbol{\lambda_1},\boldsymbol{\lambda_2}\dots$ such that
\begin{equation*}
\displaystyle\lim_{t\rightarrow \infty}
B_F(0\|\mathcal{L}_F^{\boldsymbol{\lambda}_t})
  \rightarrow  \arg \displaystyle\min_{\boldsymbol{\lambda}\in \mathbb{R}^n} B_F(0\|\mathcal{L}_F^{\boldsymbol{\lambda}})
\end{equation*}
subject to
\begin{equation*}
\|\boldsymbol{\lambda}\|_1\le \boldsymbol{u}
\end{equation*}

 Let $\boldsymbol{\lambda}_1=\mathbf{0}$

\textbf{For} $k=1,2,\dots$

$~~\boldsymbol{q}^k=\mathcal{L}_F^{\boldsymbol{\lambda}_k}$

$~~\boldsymbol{\delta}_k = \arg \displaystyle\min_{\boldsymbol{\delta}\in \mathbb{R}^n}
  \sum_{i=1}^m q_i^k(e^{-(\boldsymbol{\lambda}^TA)_i}-1)$

$~~~~st: \;\|\;\boldsymbol{\lambda}_k + \boldsymbol{\delta}_k\;\|\;_1\le \boldsymbol{u}$


~~Update
$~~\boldsymbol{\lambda}_{k+1}=\boldsymbol{\lambda}_k +
\boldsymbol{\delta}_k$

\textbf{End For}

\section{A Primal-Dual method for $L1$ regularized Logistic Regression}
The basic algorithm for the unconstrained case was proposed
by~\cite{Collins02}, but their method finds a lower bound
using the first order characteristics of the unconstrained
minimizer. In our case we want to find the constrained minimizer of
the auxiliary function. Since we need strict non-negative
$\mathcal{A}(\boldsymbol{\delta},\boldsymbol{q})\le 0$, so the new
set of conditions are
\begin{eqnarray}\label{EQN:ConstrainedMinimization_1}
  \arg \displaystyle\min_{\boldsymbol{\delta}\in \mathbb{R}^n}&&
  \sum_{i=1}^m q_i(e^{-(\boldsymbol{\delta}^TA)_i}-1)  \\ \nonumber
  st:&& \;\|\;\boldsymbol{\lambda} + \boldsymbol{\delta}\;\|\;_1\le u
  \\ \nonumber
  &&\mathcal{A}(\boldsymbol{\delta},\boldsymbol{q})\le 0
\end{eqnarray}

Analyzing the cost function more closely we find that it can be
written as
\begin{eqnarray*}
  e^{-(\boldsymbol{\delta}^TA)_i} -1&=& e^{-\sum_{j=1}^n(\boldsymbol{\delta}_jA_{ji})_i}
  -1 \\
   &=& e^{-\sum_{j=1}^n(\boldsymbol{\delta}_js_{ji}|A_{ji}|)}
   -1\hspace{1.0cm} \\
 &\le& \sum_{j=1}^n|A_{ji}|(e^{-(\boldsymbol{\delta}_js_{ji})}
   -1)
\end{eqnarray*}
where $s_{ji}=sign(A_{ji})$. Absorbing, this constraint into the
cost function we get
\begin{eqnarray}\label{EQN:ConstrainedMinimization_2}
  \arg \displaystyle\min_{\boldsymbol{\delta}\in \mathbb{R}^n}&&
  \sum_{i=1}^m q_i\sum_{j=1}^n|A_{ji}|(e^{-(\boldsymbol{\delta}_js_{ji})}
   -1)	\\ \nonumber
  st:&& \;\|\;\boldsymbol{\lambda} + \boldsymbol{\delta}\;\|\;_1\le
  \boldsymbol{u}\\ \nonumber
   &&\mathcal{A}(\boldsymbol{\delta},\boldsymbol{q})\le 0
\end{eqnarray}
Now we define the two quantities
\begin{eqnarray*}
  W_j^+(\boldsymbol{q}) &=& \displaystyle\sum_{sign(A_{ji})=+1} q_i|A_{ji}|\\
  W_j^-(\boldsymbol{q}) &=& \displaystyle\sum_{sign(A_{ji})=-1} q_i|A_{ji}|\\
\end{eqnarray*}
such that at iteration $k$ we have $W_j^+(\boldsymbol{q}_t)$ and
$W_j^-(\boldsymbol{q}_t)$, then we can re-write the optimization
problem as
\begin{equation*}
  \arg \displaystyle\min_{\boldsymbol{\delta}\in \mathbb{R}^n}
  \sum_{j=1}^nW_j^+(\boldsymbol{q}_t)(e^{-\boldsymbol{\delta}_j}-1)+
  W_j^-(\boldsymbol{q}_t)(e^{\boldsymbol{\delta}_j}-1)
\end{equation*}
\begin{eqnarray}\label{EQN:ConstrainedMinimization_3}
  st:&& \|\boldsymbol{\lambda} + \boldsymbol{\delta}\|_1\le
  \boldsymbol{u}\\ \nonumber
   &&\mathcal{A}(\boldsymbol{\delta},\boldsymbol{q})\le 0
\end{eqnarray}
Adopting from \cite{Victor01}, we can now introduce slack
variables and write the penalty function as
\begin{eqnarray}\label{EQN:ConstrainedMinimization_4}
  \nonumber \arg \displaystyle\min_{\boldsymbol{\delta},\boldsymbol{r},\boldsymbol{s},\boldsymbol{t},\boldsymbol{u}\in \mathbb{R}^n}&&
  \sum_{j=1}^n\mathcal{G}(\delta_j)+ a\mathbf{e}^T(s_j+t_j)
    \\
  st:&& \lambda_j + \delta_j + s_j - t_j=u_j \\ \nonumber
     && \mathcal{G}(\delta_j)+r_j=0\\ \nonumber
     && s_j, t_j, r_j \ge 0
\end{eqnarray}

where
$\mathcal{G}(\delta_j)=W_j^+(\boldsymbol{q}_t)(e^{-\boldsymbol{\delta}_j}-1)+
  W_j^-(\boldsymbol{q}_t)(e^{\boldsymbol{\delta}_j}-1)$ and $j=\{1,\dots,n\}$.

Finally, introducing the log barrier function and absorbing the two
terms $\lambda_j$ and $u_j$ into one term $c_j=u_j-\lambda_j$ we get
\begin{equation*}\label{EQN:ConstrainedMinimization_5}
  \arg \displaystyle\min_{\boldsymbol{\delta},\boldsymbol{r},\boldsymbol{s},\boldsymbol{t},\boldsymbol{c}\in \mathbb{R}^n}
  \sum_{j=1}^n\mathcal{G}(\delta_j)+ a\mathbf{e}^T(s_j+t_j) -\mu\boldsymbol{\phi}(s_j, t_j, r_j)
\end{equation*}
\begin{eqnarray}\label{EQN:ConstrainedMinimization_5}
  st:&& \delta_j + s_j - t_j=c_j\\ \nonumber
  &&\mathcal{G}(\delta_j) + r_j = 0\\ \nonumber
\end{eqnarray}
where $\boldsymbol{\phi}(s_j, t_j,
r_j)=\log{s_j}+\log{t_j}+\log{r_j}$ and $\mu$ is the barrier
parameter. As proposed in \cite{Victor01}, we decompose the
problem into a master problem and a sequence of sub-problems. We
solve the following master problem for a sequence of barrier
parameters $\{\mu_k\}$ such that
$\displaystyle\lim_{k\rightarrow\infty}\mu_k=0+$ where the $+$ sign
denotes converging to $0$ from the positive side
\begin{equation*}
\displaystyle\min_c\sum_{j=1}^NF_j^{\star}(\mu,c)
\end{equation*}

The sequence of subproblems are exactly same as
Eqn.~\ref{EQN:ConstrainedMinimization_5}, except the fact that the
value of $c$ is held constant while solving the sub-problems. The
$j^{th}$ sub-problem can now be written as
\begin{eqnarray}\label{EQN:ConstrainedMinimization_6}
  \arg \displaystyle\min_{{\delta},{r},{s},{t}\in \mathbb{R}}&&
  \mathcal{G}(\delta)+ a(s+t) -\mu\boldsymbol{\phi}(s, t, r)\\ \nonumber
  st:&& \delta + s - t=c \\ \nonumber
  &&\mathcal{G}(\delta) + r = 0\\ \nonumber
\end{eqnarray}
Proceeding as shown in Convex Optimization~\cite{Boyd04}, Eqn.
$11.53$, the modified KKT conditions can be expressed as
$\mathfrak{r}_t(x,\lambda,\nu)=0$, (where the $(\lambda,\nu)$ are
the multipliers, redefined again for consistency of notation), where
we define
\begin{equation}\label{Eqn:KKT}
\mathfrak{r}_t(x,\lambda,\nu)=\left[
                               \begin{array}{c}
                                 \nabla f_0(x)+ J(x)^T\lambda + A^T\nu\\
                                 (\lambda)f(x)-\mu \\
                                 Ax-b \\
                               \end{array}
                             \right]=0
\end{equation}
where
\begin{eqnarray*}
  x &=& \left[
          \begin{array}{c}
            \delta,~ r,~ s, ~t \\
          \end{array}
        \right]^T \\
  f_0(x) &=& \mathcal{G}(\delta)+ a(s+t) -\mu\boldsymbol{\phi}(s, t, r) \\
  f(x) &=& \mathcal{G}(\delta) + r \\
  J(x) &=& \left[
             \begin{array}{c}
               \Delta \mathcal{G}(\delta),\; 1, \;0, \;0\\
             \end{array}
           \right]^T \\
  A &=& \left[
          \begin{array}{c}
            1,\; 0,\; 1,\; -1\\
            \end{array}
        \right]^T
   \\
  b &=& c
\end{eqnarray*}
The Newton step can be now be formulated as
\begin{eqnarray}\label{Eqn:NewtonStep}
\left[
  \begin{array}{ccc}
    {\nabla^2f_0(x)+\lambda \nabla^2f(x)} & {J(x)^T} & {A^T} \\
    {\lambda J(x)} & {f(x)} & {0} \\
    {A} & 0 & 0 \\
  \end{array}
\right]\cdot \\ \nonumber \left[
         \begin{array}{c}
           \nabla x \\
           \nabla \lambda \\
           \nabla \nu \\
         \end{array}
       \right]=-\left[
                  \begin{array}{c}
                    \mathfrak{r}_{dual} \\
                    \mathfrak{r}_{cent} \\
                    \mathfrak{r}_{pri} \\
                  \end{array}
                \right]&&
\end{eqnarray}
where
\begin{equation*}
\left[
                  \begin{array}{c}
                    \mathfrak{r}_{dual} \\
                    \mathfrak{r}_{cent} \\
                    \mathfrak{r}_{pri} \\
                  \end{array}
                \right]=\mathfrak{r}_t(x,\lambda,\nu)
\end{equation*}

\section{Experiments and Results}
In this section we report results for the experiments conducted for
the new model proposed in this paper. The sparsity introduced by the
$L1$ regularization is captured by conducting tests on randomly
generated data. The loss-minimization curves remain similar to the
unconstrained case since the unit slave problems mentioned in
Eqn.~\ref{EQN:ConstrainedMinimization_5} are convex. But the
sparsity of feature vectors enables the dropping of redundant
features and hence speeds up the iterations.
\begin{figure*}[ht!]
\begin{center}
\includegraphics[width=8cm]{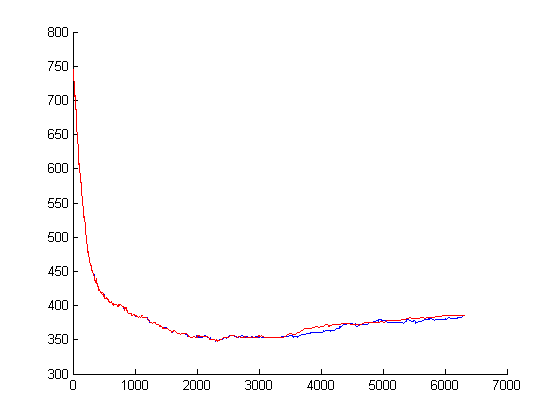}
\includegraphics[width=8cm]{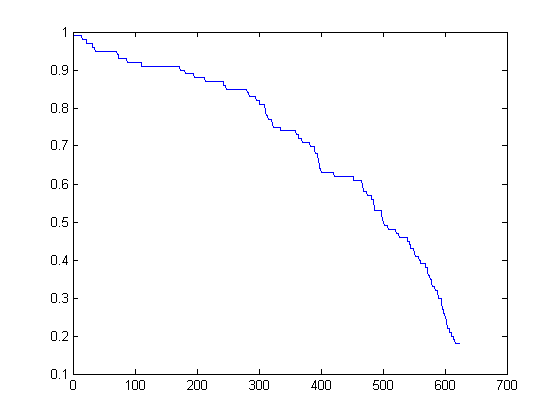}
\caption{Left: Test Error, regularized (blue) and unconstrained
(red) for $500$D, Right: Dropped features as a percentage of the
total features.} \label{Fig:randomRes}
\end{center}
\end{figure*}

In our experiments, we generated random data and classified it using
a very noisy hyperplane. We investigate only $2$-class
classification problems in this work. We investigate medium to high
dimensional problems where the dimensionality ranges from $20-500$.
We tested both the scenarios a) when the number of training points
is of the order of the feature dimension and b) when the number of
the training data points is an more than an order from the feature
dimension. For every case the random data is first classified based
on a random hyperplane and then we add Gaussian noise to the data
dimensions based on a coin flip. The noise is assumed to be
$\epsilon\sim \mathcal{N}(\mathbf{0},\sigma \mathbf{I})$, where
$\sigma<1$. The key point of interest is the fact that since the
procedure mentioned in this work decouples the features, and hence
the features are dropped from the optimization scheme when the
change $\nabla \delta_i$ drops below some threshold. One such
comparative plots are shown in Fig.~\ref{Fig:randomRes} (left). The
sparsity of feature is shown in Fig.~\ref{Fig:randomRes} (right).

\begin{figure*}[ht!]
\begin{center}
\includegraphics[width=8cm]{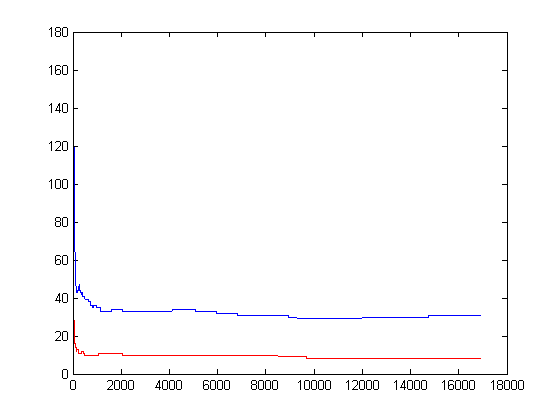}
\includegraphics[width=8cm]{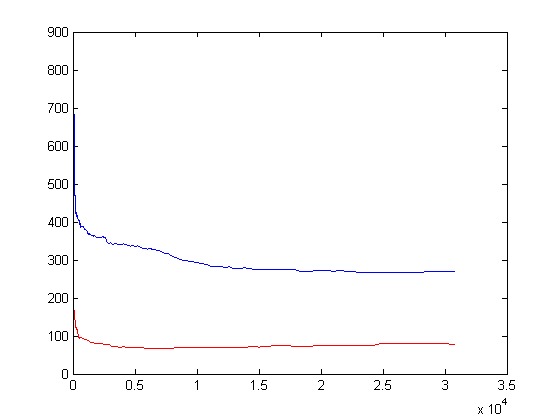}
\caption{Train Error (blue) and Test Error (red). Left: WDBC
data, Right: Musk Clean
$2$ data.} \label{Fig:WDBCRes}
\end{center}
\end{figure*}

For comparing with other algorithms we run the logistic classifier
over public domain data namely the Wisconsin Diagnostic Breast
Cancer (WDBC) data set and the Musk data base (Clean $1$ and
$2$)~\cite{Newman98}. The WDBC data has $569$ instances with $30$
real valued features. There are 357 benign (positive) instances and
$212$ malignant (negative) instances. The best reported result is
$97.5\%$ using decision trees constructed by linear
programming~\cite{Mangasarian95,Bennett92}. Our method generate $16$
fakse negatives and $23$ false positives, totaling $39$ errors with
an accuracy of $93.15\%$. The training and testing errors are shown
in Fig.~\ref{Fig:WDBCRes} (left).

The musk clean $1$ data-set describes a set of 92 molecules of which
47 are judged by human experts to be musks and the remaining 45
molecules are judged to be non-musks. Similarly, the musk clean $2$
data base describes a set of $102$ molecules of which $39$ are musks
and the remaining $63$ molecules are non-musks. The $166$ features
that describe these molecules depend upon the exact shape, or
conformation, of the molecule. Multiple confirmations for each
instance were created, which after pruning amount to $476$
conformations for clean $1$ and $6598$ for clean $2$ data-set. The
many-to-one relationship between feature vectors and molecules is
called the "multiple instance problem". When learning a classifier
for this data, the classifier should classify a molecule as "musk"
if ANY of its conformations is classified as a musk. A molecule
should be classified as "non-musk" if NONE of its conformations is
classified as a musk.

We report results for tests conducted on the two data-bases. The
training and test plots for the clean $2$ data are shown in
Fig.~\ref{Fig:WDBCRes} (right). We compare our method L$1$ Logistic
Regression based on Bregman Distances (L1LRB) against published
results and our method outperforms most of them. The comparative
results are shown in Table.~\ref{Tab:MuskClean1Res} and
Table.~\ref{Tab:MuskClean2Res}. Also note that the poor performance
of C$4.5$ algorithm has been attributed to the fact that it does not
take the multi-instance nature of the problem into consideration for
training. We did not take this consideration while training and
still our method ranks as the top $2$ for among all the reported
results. The details for the other methods mentioned have been
discussed in~\cite{Dietterich97}.


\begin{table}[h]
\centering
\begin{tabular}{|c|c|c|c|c|c|}
  \hline
  Algorithm & TP & FN & FP & TN & \% Acc \\
  \hline
  L1LRB & $45$ & $2$ & $2$ & $43$ & $95.6$ \\
  Iter-discrim APR & $42$ & $5$ & $2$ & $43$ & $92.4$ \\
  GFS-Elim-kde APR & $46$ & $1$ & $7$ & $38$  & $91.3$ \\
  All-pos APR & $36$ & $11$ & $7$ & $38$  & $80.4$ \\
  Back-prop & $45$ & $2$ & $21$ & $24$	& $75.0$ \\
  C4.5(pruned) & $45$ & $2$ & $24$ & $21$  & $68.5$ \\
  \hline
\end{tabular}
\caption{Comparative results for the Musk Clean $1$
database.}\label{Tab:MuskClean1Res}
\end{table}

\begin{table}[h]
\centering
\begin{tabular}{|c|c|c|c|c|c|}
  \hline
  Algorithm & TP & FN & FP & TN & \% Acc \\
  \hline
  Iter-discrim APR & $30$ & $9$ & $2$ & $61$ & $89.2$ \\
  L1LRB & $30$ & $9$ & $6$ & $57$ & $85.29$ \\
  GFS-Elim-kde APR & $32$ & $7$ & $13$ & $50$  & $80.4$ \\
  GFS-El-count APR & $31$ & $8$ & $17$ & $46$  & $75.5$ \\
  All-pos APR & $34$ & $5$ & $23$ & $40$  & $72.6$ \\
  Back-prop & $16$ & $23$ & $10$ & $53$	 & $67.7$ \\
  GFS-All-Pos APR & $37$ & $2$ & $32$ & $31$  & $66.7$ \\
  Most Freq Class & $0$ & $39$ & $0$ & $63$	 & $61.8$ \\
  C4.5(pruned) & $32$ & $7$ & $35$ & $28$  & $58.8$ \\
  \hline
\end{tabular}
\caption{Comparative results for the Musk Clean $2$
database.}\label{Tab:MuskClean2Res}
\end{table}

\section{Conclusion and extensions}We posed the problem of $L1$
regularized logistic regression as a constrained Bregman distance
minimization problem and posed the optimization problem as a
decoupled primal-dual problem in each of the dimensions of the
parameter vector. The optimization technique mentioned in this work
takes help from the strict feasibility properties of primal dual
methods and hence guarantee the convergence of the algorithm.
Comparative results on published data-sets have prove the strength
of the regularized method.



\bibliographystyle{latex8}
\bibliography{project510}

\end{document}